\documentclass[10pt,twocolumn,letterpaper]{article}

\usepackage[usepackage]{cvpr}      % To produce the REVIEW version
\usepackage{cuted} %< for strip environment

\usepackage{currfile} %< for \currfilebase macro

\usepackage{caption} %< for "\captionof" commands (teaser)
\usepackage{graphicx}
\usepackage{cuted}   % 提供跨双栏 strip 环境
\usepackage{caption} % 提供 \captiono
\usepackage{booktabs}
\usepackage{array}
\usepackage{float}
\usepackage{cuted}
\usepackage{capt-of}
\definecolor{cvprblue}{rgb}{0.21,0.49,0.74}
\usepackage[pagebackref,breaklinks,colorlinks,allcolors=cvprblue]{hyperref}

\def\paperID{*****} % *** Enter the Paper ID here
\def\confName{CVPR}
\def\confYear{2026}

\title{GCSR-OCC: Geometry Calibration and Semantic Routing for Open-Vocabulary Occupancy Prediction}

\author{
Bohua Wang\thanks{Equal contribution.}
\quad
Congyang Zhao\footnotemark[1]
\quad
Tianxiong Lv\thanks{Corresponding author.} \quad
Chenxu Zhao \quad 
He Zheng \quad 
Hangqi Fan \quad \\
Leigang Luo \quad
Zhihao Liao \quad
Yihong Guo \quad
Yang Cai\\
Alibaba Amap\\
{\small\texttt{wangbohua.wbh}}, {\small\texttt{zhaocongyang.zcy@alibaba-inc.com}}\\
{\small\texttt{tianxiong.ltx@autonavi.com}}
}
\usepackage{etoolbox}
\begin{document}
\maketitle
\setlength{\stripsep}{0pt}
\setlength{\stripsep}{0pt}
\begin{strip}
\centering
\includegraphics[width=\linewidth]{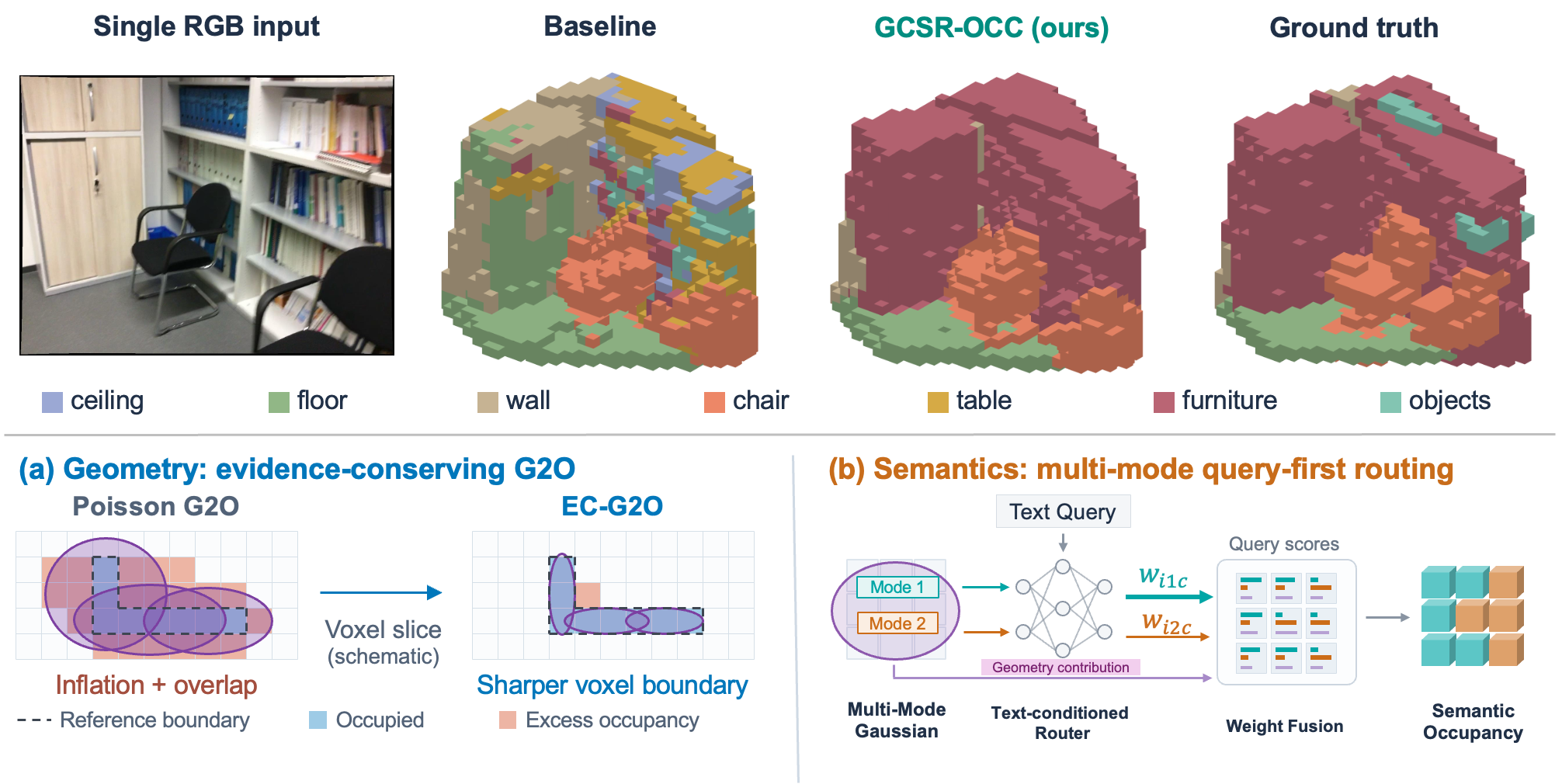}
\captionsetup{hypcap=false}
\captionof{figure}{
    \textbf{Overview of GCSR-OCC.}
    Our method predicts geometric occupancy and open-vocabulary
    voxel semantics from a single RGB image.
    Top: a selected qualitative comparison with a reproduced
    LegoOcc baseline and ground truth.
    Bottom: geometry calibration, query-first semantic routing,
    and reported performance.
}
\label{fig:teaser}
\captionsetup{hypcap=true}
\end{strip}
\begin{abstract}
Open-vocabulary 3D occupancy prediction represents scene geometry and semantics in a unified voxel grid, enabling recognition beyond predefined categories. One approach reconstructs scenes with 3D Gaussians and converts them into voxel occupancy. However, Gaussian scale inflation and excessive overlap can distort voxel occupancy predictions, while semantic mixing can arise within Gaussians spanning multiple categories and during 3D--2D semantic feature alignment. We propose GCSR-OCC, which combines geometry calibration and multi-mode query-conditioned semantic routing for indoor occupancy prediction from a single RGB image. The Evidence-Conserving Gaussian-to-Occupancy operator (EC-G2O) uses kernel mass normalization, $p$-norm aggregation, and a learnable gain to suppress scale inflation and excessive local evidence accumulation while calibrating occupancy responses. Multi-mode Query-First Gaussian Semantic Routing (MM-QGSR) matches multiple semantic modes per Gaussian to text queries and aggregates query-specific evidence into voxels through a learnable text-conditioned router. Fine-grained subclass queries and smoothed semantic aggregation further refine voxel-level predictions. Without 3D semantic annotations, GCSR-OCC achieves state-of-the-art performance in open-vocabulary occupancy prediction on Occ-ScanNet, with 63.31\% geometric IoU and 27.44\% semantic mIoU. It also reduces geometric aggregation latency from $19.76\,\mathrm{ms}$ to $12.88\,\mathrm{ms}$. Our code will be made publicly available.
\end{abstract}
\section{Introduction}
\label{sec:intro}

Semantic 3D occupancy represents scene geometry and semantics in a unified voxel grid~\cite{song2017sscnet,tian2023occ3d,wang2023openoccupancy}, supporting robot navigation and interaction with the environment~\cite{hornung2013octomap,huang2023vlmaps}. Real-world categories are diverse and difficult to enumerate, motivating semantic recognition beyond a predefined set of semantic categories~\cite{gu2022vild,takmaz2023openmask3d}. This paper addresses monocular open-vocabulary 3D occupancy prediction in indoor scenes, aiming to predict voxel occupancy states and open-vocabulary semantics from a single RGB image.

Direct 3D occupancy prediction from a single RGB image faces challenges from depth ambiguity and occlusion of scene regions by foreground objects, while dense voxel representations entail a trade-off between spatial resolution and computational cost~\cite{tatarchenko2017ogn,roldao2020lmscnet}. An alternative is to first reconstruct the scene in 3D and then convert the reconstruction into voxel occupancy, a paradigm we term Reconstruction-then-Occupancy (RTO).

RTO uses reconstructed 3D structure to guide occupancy prediction. It can leverage geometric priors from pretrained visual geometry models~\cite{yang2024depthanythingv2,wang2024dust3r,wang2025vggt,eftekhar2021omnidata} and incorporate auxiliary supervision on the reconstructed scene to improve geometric consistency~\cite{huang2024gaussianformer,huang2025gaussianformer2,zhou2026legoocc,yu2022monosdf}. However, 3D occupancy prediction accuracy depends not only on reconstruction quality but also on the reconstruction-to-occupancy conversion mechanism, which is the focus of this work.

Direct alignment between 3D representations and text remains challenging due to the scarcity of large-scale 3D--text datasets~\cite{xue2023ulip,qin2024langsplat}. Therefore, several existing methods use pretrained 2D vision--language models~\cite{radford2021clip,ghiasi2022openseg} for semantic supervision. These methods transfer language-aligned features to 3D through image-space correspondences~\cite{jatavallabhula2023conceptfusion} or rendering supervision~\cite{kerr2023lerf}, without requiring 3D semantic annotations~\cite{boeder2024langocc,vobecky2023pop3d,zhou2026legoocc}. However, 3D-to-2D mapping can associate multiple 3D locations with the same image pixel~\cite{wu2024opengaussian}. Aggregating their features can mix distinct semantics and degrade voxel-level semantic predictions.

LegoOcc~\cite{zhou2026legoocc} is a representative reconstruction-based approach to monocular open-vocabulary 3D occupancy prediction in indoor scenes. Its opacity-aware, Poisson-based Gaussian-to-occupancy operator stabilizes volumetric aggregation under weak supervision. However, this formulation allows the model to increase occupancy responses through Gaussian scale inflation and greater overlap, without necessarily improving 3D geometric accuracy. On the semantic side, it aggregates Gaussian features before text matching, potentially merging different semantics into a single voxel feature. Although temperature scheduling reduces feature blending during training, text queries still operate on this aggregated representation.

To address these limitations, we propose GCSR-OCC (Figure 2), which combines geometry calibration and query-conditioned semantic routing for Gaussian-to-voxel aggregation. Our main contributions are as follows:

\begin{itemize}
    \item We introduce an \textbf{Evidence-Conserving Gaussian-to-Occupancy operator (EC-G2O)}, combining kernel mass normalization, $p$-norm aggregation, and a learnable gain to suppress scale inflation and excessive local evidence accumulation while calibrating occupancy responses.

    \item We propose \textbf{Multi-mode Query-First Gaussian Semantic Routing (MM-QGSR)}, which improves semantic prediction through mode-level text matching and text-conditioned routing. Subclass queries and smoothed aggregation further refine voxel-level semantics.

    \item On Occ-ScanNet, GCSR-OCC achieves $63.31\%$ geometric IoU and $27.44\%$ open-vocabulary semantic mIoU. Compared with the Poisson baseline, geometric aggregation latency decreases from $19.76\,\mathrm{ms}$ to $12.88\,\mathrm{ms}$.
\end{itemize}
\begin{figure*}[t]
    \centering
    \includegraphics[width=\textwidth]{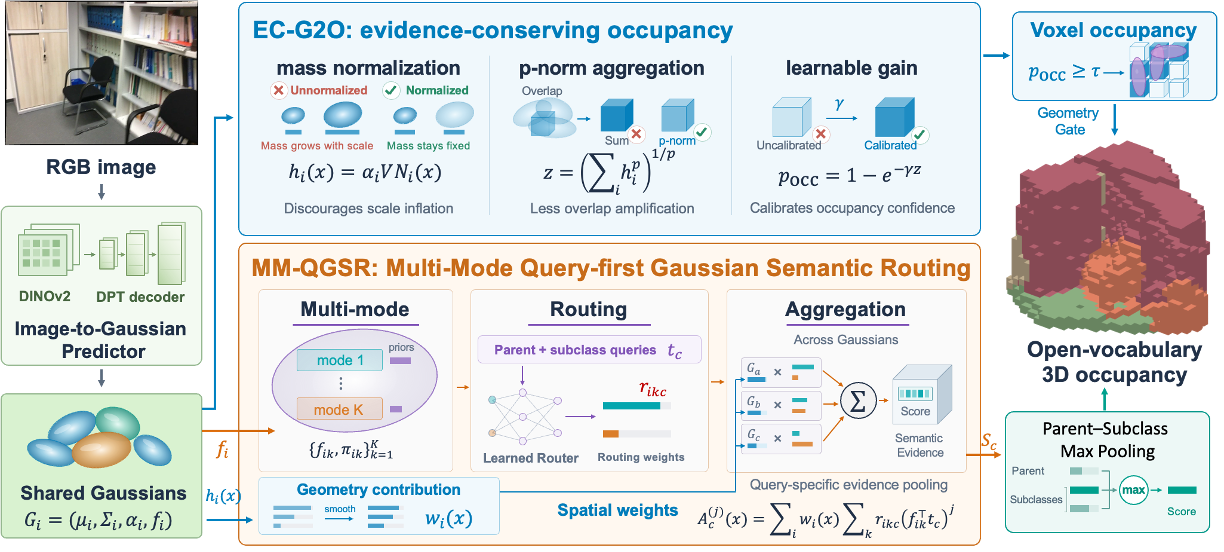}
    \caption{Overview of the proposed GCSR-OCC framework.}
    \label{fig:architecture}
\end{figure*} 

\section{Related Work}
\label{sec:related_work}

\subsection{3D Semantic Occupancy Prediction}
\label{sec:related_work_occupancy}

Vision-based occupancy prediction lifts image information into 3D to recover scene geometry and semantics~\cite{wei2023surroundocc,zhang2023occformer}. MonoScene~\cite{cao2022monoscene} combines the Features Line of Sight Projection (FLoSP) module with a 3D context prior for monocular scene completion. VoxFormer~\cite{li2023voxformer} initializes sparse voxel queries from estimated depth and propagates their information to dense voxels through attention. TPVFormer~\cite{huang2023tpvformer} uses three orthogonal planes to lift and aggregate image features in 3D, enabling dense semantic occupancy prediction under sparse supervision. For indoor scenes, ISO~\cite{yu2025iso} combines a pretrained depth model with dual feature projection and introduces Occ-ScanNet. EmbodiedOcc~\cite{wu2025embodiedocc} maintains a Gaussian scene memory across successive observations for online occupancy prediction. This paper follows a reconstruction-then-occupancy framework and focuses on converting reconstructed 3D representations into voxel occupancy.

\subsection{Reconstruction-Based Occupancy Prediction}
\label{sec:related_work_reconstruction}

Reconstruction-based methods model scene geometry using implicit fields~\cite{mescheder2019occupancy,park2019deepsdf,mildenhall2020nerf,oechsle2021unisurf,wang2021neus} or explicit 3D primitives~\cite{paschalidou2019superquadrics,kerbl20233d,guedon2024sugar,huang20242dgs}, providing spatial constraints for occupancy prediction. SelfOcc~\cite{huang2024selfocc} represents scenes as signed distance fields and learns 3D occupancy through cross-frame image reconstruction. GaussianOcc~\cite{gan2025gaussianocc} learns metric-scale 3D occupancy through Gaussian splatting, using cross-camera constraints and efficient voxel rendering.

Another line of work directly aggregates Gaussian primitives into voxel occupancy. GaussianFormer~\cite{huang2024gaussianformer} reduces memory costs with sparse semantic Gaussians and converts them into dense voxel occupancy through splatting. GaussianFormer-2~\cite{huang2025gaussianformer2} models geometric occupancy through probabilistic superposition and semantics through a Gaussian mixture model. EmbodiedOcc++~\cite{wang2025embodiedoccpp} improves Gaussian representations through planar regularization and uncertainty-guided updates. RoboOcc~\cite{zhang2025roboocc} uses opacity-guided interactions and geometry-aware encoding to improve semantic and geometric prediction. Our EC-G2O operator focuses on Gaussian-to-voxel aggregation, suppressing scale inflation and excessive local evidence accumulation.

\subsection{Open-Vocabulary Occupancy Prediction}
\label{sec:related_work_open_vocabulary}

Language-aligned 2D~\cite{radford2021clip,li2022lseg,zhou2022maskclip} and 3D representations~\cite{peng2023openscene} support open-vocabulary scene understanding. POP-3D~\cite{vobecky2023pop3d} distills 2D vision--language features into voxels for text-based 3D recognition. LangOcc~\cite{boeder2024langocc} renders 3D language features into image space and aligns them with 2D vision--language features. These methods support open-vocabulary occupancy prediction without 3D semantic annotations.

Gaussian-based methods adopt different paradigms for aggregating semantic information into voxels. LegoOcc~\cite{zhou2026legoocc} aggregates Gaussian features into voxel features before text matching, using progressive temperature decay to reduce feature blending during rendering supervision. In contrast, GaussTR~\cite{jiang2025gausstr} performs text matching on individual Gaussians and aggregates the resulting semantic scores into voxels. Feature aggregation before text matching can mix distinct semantics. Although GaussTR matches text before aggregation, each Gaussian carries a single semantic feature, and the aggregation weights are not explicitly text-conditioned. To address these limitations, our MM-QGSR represents each Gaussian with multiple semantic modes and aggregates query-specific evidence into voxels through learnable text-conditioned routing.
\section{Method}
\label{sec:method}
% Problem definition. Included from sec/3_method.tex, before the geometry subsection.
\subsection{Problem Definition}
\label{sec:problem}

The input is a monocular RGB image of an indoor scene, $\mathbf{I}\in\mathbb{R}^{H\times W\times 3}$. Predictions are made on a regular voxel grid in front of the camera, with resolution $X\times Y\times Z$ and voxel volume $V$. At a voxel center $\mathbf{x}$, the model predicts an occupancy probability $p_{\mathrm{occ}}(\mathbf{x})$ and a category if the voxel is occupied. Categories are not restricted to a closed list fixed at training time. At inference, the category is obtained by matching a text query to the semantic feature of that voxel.

The two predictions use one set of 3D primitives. A decoder maps $\mathbf{I}$ to anisotropic 3D Gaussians $\mathcal{G}=\{G_i\}_{i=1}^{N}$,
\begin{equation}
G_i=(\boldsymbol{\mu}_i,\boldsymbol{\Sigma}_i,\alpha_i,\mathbf{f}_i).
\end{equation}
$\boldsymbol{\mu}_i\in\mathbb{R}^3$ and $\boldsymbol{\Sigma}_i$ specify position and shape. The covariance factors as $\boldsymbol{\Sigma}_i=R_i^{\top}S_i^{2}R_i$, with rotation $R_i$ and semi-axes $S_i=\mathrm{diag}(s_{i1},s_{i2},s_{i3})$. $\alpha_i\in(0,1]$ is the opacity, and $\mathbf{f}_i$ is a semantic feature. The occupancy probability depends only on $(\boldsymbol{\mu}_i,\boldsymbol{\Sigma}_i,\alpha_i)$; $\mathbf{f}_i$ does not enter $p_{\mathrm{occ}}$. The same $\alpha_i$ is also used to render $\mathcal{G}$ onto the image plane, so occupancy aggregation and rendering cannot each introduce an opacity of their own. At a query point $\mathbf{x}$, the spatial influence of $G_i$ is the unnormalized kernel
\begin{equation}
\varphi_i(\mathbf{x})
=\exp\Big(-\tfrac12(\mathbf{x}-\boldsymbol{\mu}_i)^{\top}\boldsymbol{\Sigma}_i^{-1}(\mathbf{x}-\boldsymbol{\mu}_i)\Big).
\end{equation}
Since $\varphi_i(\boldsymbol{\mu}_i)=1$ and the expression excludes $\lvert\boldsymbol{\Sigma}_i\rvert$, changing the semi-axes changes coverage without changing the peak. The next subsection defines the aggregator that produces $p_{\mathrm{occ}}$ from $\mathcal{G}$; category prediction is described with the semantic model.

% Geometry subsection. Included from sec/3_method.tex.
\subsection{Evidence-Conserving Gaussian-to-Occupancy}
\label{sec:ecg2o}

We aggregate the decoder's 3D Gaussians into voxel occupancy. The same primitives are $\alpha$-blended in image space, so the 2D and 3D pathways share a single opacity $\alpha_i$. Unlike aggregators that set the peak occupancy of a single Gaussian to $1$ (e.g. GaussianFormer-2), this shared-opacity design prevents omitting $\alpha$ in the 3D geometry or introducing a separate 3D opacity.

Previous Poisson-based Gaussian-to-Occupancy (G2O) retains the shared $\alpha$, but uses the unnormalized splatting kernel as an intensity. A single primitive cannot yield high occupancy confidence. Because a linear sum treats each overlapping primitive as an independent hit, merely duplicating a Gaussian increases occupancy just as effectively as placing it accurately. Consequently, training exploits a shortcut: compensating for inaccurate placement and weak per-primitive confidence by generating more and larger Gaussians. We propose evidence-conserving G2O (EC-G2O). Evidence is a Gaussian's contribution to occupancy; conservation means that the integrated contribution does not grow with scale. Given the shared $\alpha$, EC-G2O decouples the integrated mass from the spatial scale and attenuates the extra occupancy contributed by overlapping primitives. 

% Given the shared $\alpha$, EC-G2O decouples the integrated mass from the spatial scale, preventing overlapping regions from being scored as independent hits. We illustrate this comparison in Figure~\ref{fig:geo}.
\begin{figure}[t]
    \centering
    \includegraphics[width=\columnwidth]{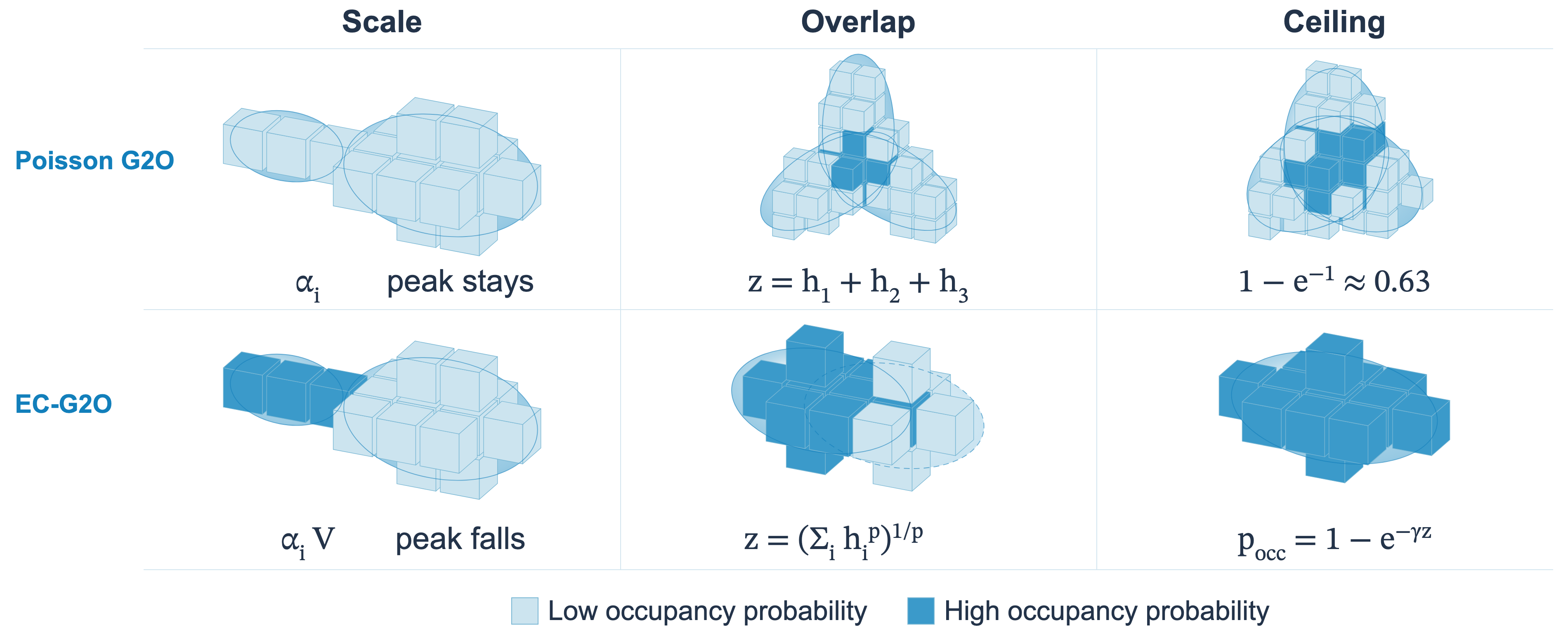}
    \caption{Geometric effects of EC-G2O (darker blocks indicate higher occupancy probability).
    Top (Scale and Overlap): Under Poisson aggregation, enlarged Gaussians expand spatial coverage without peak penalty, and overlapping regions accumulate linearly.
    Bottom (Scale and Overlap): EC-G2O dilutes enlarged Gaussians and reduces the extra occupancy caused by overlap.
    Right (Ceiling): Unlike Poisson aggregation, which requires multiple overlapping primitives for high occupancy confidence, EC-G2O allows a single Gaussian to produce a high-confidence response.}
    \label{fig:geo}
\end{figure}

\paragraph{Poisson Gaussian-to-Occupancy.}
Defining $h_i(\mathbf{x})=\alpha_i\,\varphi_i(\mathbf{x})$, Poisson aggregation computes
\begin{equation}
z(\mathbf{x})=\sum_{i=1}^{N}h_i(\mathbf{x}),
\qquad
p_{\mathrm{occ}}(\mathbf{x})=1-\exp\big(-z(\mathbf{x})\big).
\end{equation}
Here, $h_i$ is the occupancy contribution of the $i$-th Gaussian, and $z$ the accumulated contribution at the query point $\mathbf{x}$. This formulation is closed-form and naturally retains the shared opacity $\alpha_i$.

\paragraph{Limitations of Poisson G2O.}
This formulation incentivizes the model to exploit a shortcut: occupancy can increase without requiring more accurate placement.

First, scales tend to grow monotonically. The peak of $h_i$ equals $\alpha_i$ at any scale, while the integral $\int h_i\,\mathrm{d}\mathbf{x}=\alpha_i(2\pi)^{3/2}\lvert\boldsymbol{\Sigma}_i\rvert^{1/2}$ grows unboundedly with scale. Enlarging the semi-axes increases coverage at no cost to peak intensity; shrinking them only reduces coverage. Consequently, the optimization gradient encourages larger Gaussians, and the resulting overlaps are linearly accumulated as additional evidence.

Second, duplicating primitives inflates occupancy. Linear summation followed by $1-e^{-z}$ yields $p_{\mathrm{occ}}=1-\prod_i(1-p_i)$, where $p_i=1-e^{-h_i}$ is the occupancy produced by the $i$-th Gaussian alone. This formulation acts as a noisy-OR model: each Gaussian is treated as an independent source of hits, meaning any single hit suffices to occupy the voxel. As a result, merely copying a Gaussian independently inflates the occupancy probability. Although multiple Gaussians on the same surface should constitute a single piece of geometric evidence, duplication is cheaper than precise localization; thus, the network simply votes by count.

Third, a single Gaussian has a strict upper bound on its contribution. Because $\varphi_i\le 1$ and $\alpha_i\le 1$, the maximum occupancy contributed by a single primitive is strictly bounded by $1-e^{-1}\approx 0.63$. Reaching $p_{\mathrm{occ}}\approx 0.95$ requires $z\approx 3$. The first two properties make stacking inexpensive; the third makes it necessary. This ceiling is a direct consequence of sharing $\alpha_i$ with rendering. For instance, GaussianFormer-2 bypasses this issue by artificially setting the peak occupancy of a single Gaussian to $1$, allowing one primitive to saturate the voxel without stacking. However, we cannot adopt this workaround by simply rescaling the peak to $1$, as doing so would break the shared $\alpha_i$ constraint.

% Placed here so this wide table is shipped one page before the qualitative figure.
\begin{table*}[t]
\centering
\setlength{\belowcaptionskip}{6pt}
\setlength{\tabcolsep}{3pt}
\caption{Monocular occupancy on Occ-ScanNet. IoU is the geometric score. The remaining columns are per-class IoU, and mIoU is their mean over the eleven nonempty classes. Closed-vocabulary methods see full semantic labels during training. Open-vocabulary methods do not see semantic voxel labels.}
\label{tab:main}
\begin{tabular}{@{}lccccccccccccc@{}}
\toprule
Method & IoU & ceiling & floor & wall & window & chair & bed & sofa & table & TV & furniture & objects & mIoU \\
\midrule
\multicolumn{14}{c}{\textit{Closed-vocabulary (full annotation)}} \\
TPVFormer~\cite{huang2023tpvformer} & 33.39 & 6.96 & 32.97 & 14.41 & 9.10 & 24.01 & 41.49 & 45.44 & 28.61 & 10.66 & 35.37 & 25.31 & 24.94 \\
GaussianFormer~\cite{huang2024gaussianformer} & 40.91 & 20.70 & 42.00 & 23.40 & 17.40 & 27.0 & 44.30 & 44.80 & 32.70 & 15.30 & 36.70 & 25.00 & 29.93 \\
MonoScene~\cite{cao2022monoscene} & 41.60 & 15.17 & 44.71 & 22.41 & 12.55 & 26.11 & 27.03 & 35.91 & 28.32 & 6.57 & 32.16 & 19.84 & 24.62 \\
ISO~\cite{yu2025iso} & 42.16 & 19.88 & 41.88 & 22.37 & 16.98 & 29.09 & 42.43 & 42.00 & 29.60 & 10.62 & 36.36 & 24.61 & 28.71 \\
SurroundOcc~\cite{wei2023surroundocc} & 42.52 & 18.90 & 49.30 & 24.80 & 18.00 & 26.80 & 42.00 & 44.10 & 32.90 & 18.60 & 36.80 & 26.90 & 30.83 \\
EmbodiedOcc~\cite{wu2025embodiedocc} & 53.55 & 39.60 & 50.40 & 41.40 & 31.70 & 40.90 & 55.00 & 61.40 & 44.00 & 36.10 & 53.90 & 42.20 & 45.15 \\
EmbodiedOcc++~\cite{wang2025embodiedoccpp} & 54.90 & 36.40 & 53.10 & 41.80 & 34.40 & 42.90 & 57.30 & 64.10 & 45.20 & 34.80 & 54.20 & 44.10 & 46.20 \\
RoboOcc~\cite{zhang2025roboocc} & 56.48 & 45.36 & 53.49 & 44.35 & 34.81 & 43.38 & 56.93 & 63.35 & 46.35 & 36.12 & 55.48 & 44.78 & 47.76 \\
\midrule
\multicolumn{14}{c}{\textit{Open-vocabulary (geometry-only annotation)}} \\
POP-3D~\cite{vobecky2023pop3d} & 35.32 & 0.00 & 23.89 & 16.59 & 0.01 & 4.62 & 8.65 & 2.01 & 7.61 & 0.00 & 2.14 & 0.08 & 5.96 \\
LOcc~\cite{yu2025locc} & 36.70 & 0.00 & 24.40 & 17.37 & 1.83 & 9.31 & 10.73 & 23.42 & 11.29 & 0.01 & 3.15 & 0.19 & 9.25 \\
LegoOcc~\cite{zhou2026legoocc} & 59.50 & 17.81 & 41.65 & 28.52 & 18.91 & 19.32 & 33.05 & 34.92 & 19.16 & 5.36 & 5.88 & 6.94 & 21.05 \\
\textbf{Ours} & \textbf{63.31} & \textbf{35.27} & \textbf{44.40} & \textbf{31.62} & \textbf{23.37} & \textbf{21.05} & \textbf{38.87} & \textbf{40.07} & \textbf{20.04} & \textbf{8.62} & \textbf{26.92} & \textbf{11.64} & \textbf{27.44} \\
\bottomrule
\end{tabular}
\end{table*}

\paragraph{Evidence-Conserving G2O.}
EC-G2O retains the mapping of nonnegative evidence to $(0,1)$ via $1-e^{-(\cdot)}$. To address the three limitations identified above, we introduce the following modifications:

First, a mass-normalized kernel. We replace the splatting kernel with a Gaussian density scaled by the voxel volume:
\begin{equation}
\begin{aligned}
h_i(\mathbf{x})
&=\alpha_i\,V\,\mathcal{N}(\mathbf{x};\boldsymbol{\mu}_i,\boldsymbol{\Sigma}_i)\\
&=\frac{\alpha_i V}{(2\pi)^{3/2}s_{i1}s_{i2}s_{i3}}
\exp\Big(-\tfrac12(\mathbf{x}-\boldsymbol{\mu}_i)^{\top}\boldsymbol{\Sigma}_i^{-1}(\mathbf{x}-\boldsymbol{\mu}_i)\Big).
\end{aligned}
\end{equation}
The integrated mass $\int h_i\,\mathrm{d}\mathbf{x}=\alpha_i V$  is now independent of scale. Increasing the scale lowers the local peak while decreasing it raises the peak; consequently, spatial coverage and per-voxel evidence naturally trade off, driving the optimization toward more compact Gaussians.

Second, $p$-norm aggregation:
\begin{equation}
z(\mathbf{x})=\Bigl(\sum_{i=1}^{N}h_i(\mathbf{x})^{p}\Bigr)^{1/p},\qquad p\ge 1.
\end{equation}
For $p>1$, the aggregate is dominated by the largest $h_i$, and a contribution that is weak relative to $z$ receives a vanishing gradient. Equal copies are not removed: $n$ identical contributions $h$ give $z=n^{1/p}h$, so duplication still raises occupancy, but only sublinearly.
% For $p>1$, the aggregated value in overlapping regions is dominated by the largest $h_i$. Weaker contributions receive a vanishing gradient, ensuring that merely duplicating primitives no longer inflates occupancy.

Third, a learnable gain $\gamma$:
\begin{equation}
p_{\mathrm{occ}}=1-\exp(-\gamma z),
\end{equation}
where $\gamma=e^{\theta}$ is initialized at $1$. Because kernel normalization alters the magnitude of $z$, the gain $\gamma$ recalibrates it to a valid occupancy probability while compensating for the single-primitive ceiling. 
Crucially, $\gamma$ serves solely for calibration rather than acting as an independent source of geometric accuracy: increasing $\gamma$ raises the confidence of whatever evidence $z$ already contains, including evidence from stacked primitives.
% Crucially, $\gamma$ serves solely for calibration rather than acting as an independent source of geometric accuracy: without the first two modifications, simply increasing $\gamma$ would merely inflate the confidence of arbitrarily stacked primitives.

Detailed derivations, including the proof that gradients vanish for weak contributions when $p>1$, and the efficient evaluation of mass normalization and $h_i^{p}$ using standard splatting kernels, are provided in the supplementary material.

\paragraph{Geometry objective.}
Occupancy is trained with binary voxel labels inside the camera frustum; semantic voxel labels are not used. Let $y(\mathbf{x})\in\{0,1\}$ denote empty versus occupied (ignored voxels are dropped). The occupancy probability $p_{\mathrm{occ}}$ is supervised by a standard voxel occupancy loss $\mathcal{L}_{\mathrm{occ}}$, comprising focal, Lov\'asz, and scene-scale terms. Because $p_{\mathrm{occ}}=1-\exp(-\gamma z)$ saturates at large $z$, $\mathcal{L}_{\mathrm{occ}}$ alone cannot penalize surplus intensity. We therefore add two losses on the intensity $z$: a log-intensity regression $\mathcal{L}_{\log z}$ and an occupied-voxel overshoot penalty $\mathcal{L}_{\mathrm{over}}$. The geometry objective is
\begin{equation}
\mathcal{L}_{\mathrm{geo}}
=
\mathcal{L}_{\mathrm{occ}}
+\lambda_{\log z}\mathcal{L}_{\log z}
+\lambda_{\mathrm{over}}\mathcal{L}_{\mathrm{over}}.
\label{eq:geo_objective}
\end{equation}
The definitions of $\mathcal{L}_{\log z}$ and $\mathcal{L}_{\mathrm{over}}$, the expansion of $\mathcal{L}_{\mathrm{occ}}$, and the loss weights are given in the supplementary material.

% Semantic subsection. Not yet included from sec/3_method.tex.
\subsection{Multi-modal Query-Conditioned Gaussian Semantic Routing}
\label{subsec:mm_qgsr}

Aggregating Gaussian features prior to text alignment inevitably blends distinct semantics. As our semantic module, we propose Multimodal Query-Conditioned Gaussian Semantic Routing (MM-QGSR). It extracts text-aligned semantic evidence from individual Gaussians prior to voxel aggregation via a learnable router.

\paragraph{Query-first Gaussian semantic routing.}
Semantic evidence is computed for each Gaussian against the text queries, and only then aggregated into voxels. Let $\mathbf{f}_i$ and $\mathbf{t}_c$ denote the $\ell_2$-normalized semantic feature of the $i$-th Gaussian and the text feature of the $c$-th category, respectively. The raw semantic evidence $e_{ic}$, text-conditioned logit $\ell_{ic}$, and routing weight $r_{ic}$ are defined as:
\begin{equation}
\begin{aligned}
e_{ic} &= \mathbf{f}_i^\top \mathbf{t}_c, \\
\ell_{ic} &= e_{ic}
+ \tanh(\eta)\mathrm{cos\_sim}(W_g \mathbf{f}_i, W_t \mathbf{t}_c), \\
r_{ic} &= \exp\!\left[
\operatorname{clip}(\beta\ell_{ic},-L,L)
\right].
\end{aligned}
\label{eq:qgsr_routing}
\end{equation}
Here, $W_g$ and $W_t$ are learnable low-rank projections, $\eta$ controls the residual alignment term, $\beta>0$ modulates routing sharpness, and $L$ clips the logits to ensure numerical stability. In practice, we set the projection dimension to 64.

To aggregate this into the voxel grid, we combine it with the routing weight $r_{ic}$ and a normalized geometric weight $\bar g_i(\mathbf{x})$, which corresponds to the spatial contribution $w_i(\mathbf{x})$ detailed in our smoothed aggregation scheme. Specifically, we compute the query-specific evidence moments up to the second order at each voxel $\mathbf{x}$ as: 
\begin{equation}
A_c^{(j)}(\mathbf{x})
=
\sum_i \bar g_i(\mathbf{x})\,r_{ic}\,e_{ic}^{\,j},
\quad j\in\{0,1,2\}.
\label{eq:qgsr_moments}
\end{equation}
These aggregated moments are used to derive the normalized support, mean semantic match, and variance for each voxel-query pair:
\begin{equation}
\begin{aligned}
m_c(\mathbf{x}) &= A_c^{(0)}(\mathbf{x}), \\
\widetilde m_c(\mathbf{x}) &= \max(m_c(\mathbf{x}),\epsilon), \\
\mu_c(\mathbf{x}) &= \frac{A_c^{(1)}(\mathbf{x})}{\widetilde m_c(\mathbf{x})}, \\
v_c(\mathbf{x}) &=
\left[
\frac{A_c^{(2)}(\mathbf{x})}{\widetilde m_c(\mathbf{x})}
-\mu_c(\mathbf{x})^2
\right]_+.
\end{aligned}
\label{eq:qgsr_statistics}
\end{equation}
Here, $m_c$ is the geometry-normalized support for query $c$, $\mu_c$ is the average semantic match, and $v_c$ is the disagreement among the contributing evidence. The constant $\epsilon>0$ keeps the division numerically stable, and $[\cdot]_+$ clamps the negative values to zero.

These statistics form the semantic score of the voxel:
\begin{equation}
S_c(\mathbf{x})
=
\mu_c(\mathbf{x})
+
\lambda_m
\frac{\log\widetilde m_c(\mathbf{x})}{\beta}
-
\lambda_v v_c(\mathbf{x}),
\label{eq:qgsr_score}
\end{equation}
where $\lambda_m,\lambda_v\geq 0$ are global learnable calibration parameters shared across voxels and queries. For geometrically occupied voxels, the final semantic prediction is given by:
\begin{equation}
\hat y(\mathbf{x})=\operatorname*{arg\,max}_c S_c(\mathbf{x}).
\label{eq:qgsr_prediction}
\end{equation}
Ultimately, the learnable router dictates which primitive evidence contributes to each voxel, while the variance penalty ensures that only semantically consistent aggregations yield high scores.

% \paragraph{Multi-mode Gaussian semantics.}
% A single feature vector often needs to encode multiple distinct semantics within the same Gaussian primitive. We therefore extend each Gaussian to maintain $K$ latent semantic modes, allowing diverse interpretations to coexist and be aligned with text queries independently. The voxel-level evidence moments are generalized to:
% \begin{equation}
% A_c^{(j)}(\mathbf{x})
% =
% \sum_i \bar g_i(\mathbf{x})
% \sum_{k=1}^{K}
% r_{ikc}\,e_{ikc}^{\,j},
% \quad j\in\{0,1,2\},
% \label{eq:multimode_moments}
% \end{equation}
% where $e_{ikc}$ denotes the semantic evidence of the $k$-th mode in the $i$-th Gaussian for query $c$, and $r_{ikc}$ is its corresponding text-conditioned routing weight. The resulting statistics use the same support, mean, and variance as above. Setting $K=1$ naturally recovers the single-mode formulation, while $K>1$ preserves multiple semantic hypotheses per primitive, enabling the router to selectively aggregate the mode most relevant to the query.
% \paragraph{Multi-mode Gaussian semantics.}
% A single feature vector may encode multiple semantics within one Gaussian. We therefore extend each Gaussian to $K$ latent semantic modes:
% \begin{align}

\par
\noindent\textbf{Multi-mode Gaussian semantics.}\enspace A single feature vector may encode multiple semantics within one Gaussian. We therefore extend each Gaussian to $K$ latent semantic modes:
\begin{align}
\mathbf{f}_{ik} &= \operatorname{norm}\!\left(\mathbf{f}_i+\rho\tanh([F_\theta(\mathbf{f}_i)]_k)\right),
\label{eq:multimode_features} \\
\pi_{ik} &= \operatorname{softmax}_k(P_\psi(\mathbf{f}_i)),
\label{eq:multimode_priors}
\end{align}
where $F_\theta$ and $P_\psi$ are learnable MLPs generating mode residuals and priors, respectively. Here, $\operatorname{norm}$ denotes $\ell_2$ normalization, and $\rho=0.25$.

The mode-level evidence $e_{ikc}$ and logit $\ell_{ikc}$ follow Eq.~\eqref{eq:qgsr_routing} using $\mathbf{f}_{ik}$. Incorporating the mode prior gives
\begin{equation}
r_{ikc} = \pi_{ik}\exp\!\left[\operatorname{clip}(\beta\ell_{ikc},-L,L)\right].
\label{eq:multimode_routing}
\end{equation}
The voxel-level evidence moments become
\begin{equation}
A_c^{(j)}(\mathbf{x}) = \sum_i \bar g_i(\mathbf{x})\sum_{k=1}^{K} r_{ikc}e_{ikc}^{j}, \quad j\in\{0,1,2\}.
\label{eq:multimode_moments}
\end{equation}
These moments use the same support, mean, variance, and scoring definitions as above. Network architectures and initialization are provided in the supplementary material.

\paragraph{Fine-grained subclass queries.}
A single text prototype is too coarse for broad categories such as \textit{furniture} or \textit{objects}. To address this, each parent query $p$ is expanded into a set of fine-grained subclass queries $\mathcal{S}_p$, while retaining the original parent query. Let $\widetilde{\mathcal{S}}_p$ denote the retained subclass queries, and let $S_t(\mathbf{x})$ be the voxel score obtained by applying MM-QGSR to query $t$. The final parent-category score is derived via max-pooling over all associated queries:
\begin{equation}
\widehat{S}_p(\mathbf{x})
=
\max_{t \in \{p\} \cup \widetilde{\mathcal{S}}_p}
S_t(\mathbf{x}).
\label{eq:subclass_readout}
\end{equation}

\paragraph{Smoothed aggregation.}
Directly using raw occupancy support for semantic aggregation allows a few dominant Gaussians to overwhelm the contributions of others. To mitigate this, let $h_i(\mathbf{x})$ be the EC-G2O occupancy support of the $i$-th Gaussian at voxel $\mathbf{x}$. Before raising this support to a power $q$, we normalize the semantic aggregation weights:
\begin{equation}
w_i(\mathbf{x})
=
\frac{h_i(\mathbf{x})^q}{\sum_j h_j(\mathbf{x})^q},
\label{eq:smoothed_aggregation}
\end{equation}
and set $\bar g_i(\mathbf{x})=w_i(\mathbf{x})$. The exponent $q$ governs the relative contribution disparity: $q=1$ preserves their original proportions, whereas $q<1$ reduces the disparity among different Gaussians, yielding smoother semantic aggregation weights. In practice, we set $q=0.5$. Crucially, this smoothing is applied exclusively within the semantic branch, leaving the geometric occupancy predictions strictly unchanged. The router objective and the construction of the subclass queries are given in the supplementary material.

\par\noindent\textbf{Semantic objective.}\enspace We train the semantic router with
\begin{equation}
\begin{aligned}
\mathcal L_{\mathrm{router}}
&= \lambda_{\mathrm{img}}\mathcal L_{\mathrm{img}}+\lambda_{\mathrm{fit}}\mathcal L_{\mathrm{fit}} \\
&\quad +\lambda_{\mathrm{div}}\mathcal L_{\mathrm{div}}+\lambda_{\mathrm{mass}}\mathcal L_{\mathrm{mass}}.
\end{aligned}
\label{eq:mmqgsr_objective}
\end{equation}
The terms provide image distillation, footprint fitting, mode diversity, and prior regularization, respectively. Loss weights and detailed definitions are given in the supplementary material.
\section{Experiment}
\label{sec:experiment}
% Experiments. Included from sec/4_experiment.tex.
\subsection{Experimental Setup}
\label{sec:setup}
\paragraph{Dataset and Metrics.}
We evaluate monocular 3D occupancy prediction on Occ-ScanNet, where each sample pairs a single RGB image with a $60\!\times\!60\!\times\!36$ semantic occupancy grid ($4.8\,\mathrm{m}\!\times\!4.8\,\mathrm{m}\!\times\!2.88\,\mathrm{m}$ at $0.08\,\mathrm{m}$ resolution). The annotation space comprises one empty class and eleven semantic categories (ceiling, floor, wall, window, chair, bed, sofa, table, television, furniture, and \textit{objects}). The dataset is partitioned into 45,755 training and 19,764 validation frames, with all metrics evaluated on the full validation set.

During inference, our open-vocabulary model processes a single RGB image. Unlike closed-vocabulary baselines trained with semantic voxel labels, our geometry is trained with binary occupancy supervision only. The semantic router is then trained by distilling a frozen Trident teacher, without semantic voxel labels. Category names enter only as text queries, and semantic voxel labels are excluded from both training and evaluation.

The \textbf{implementation details} are provided in the supplementary material.

\begin{figure*}[t]
\centering
\includegraphics[width=\textwidth]{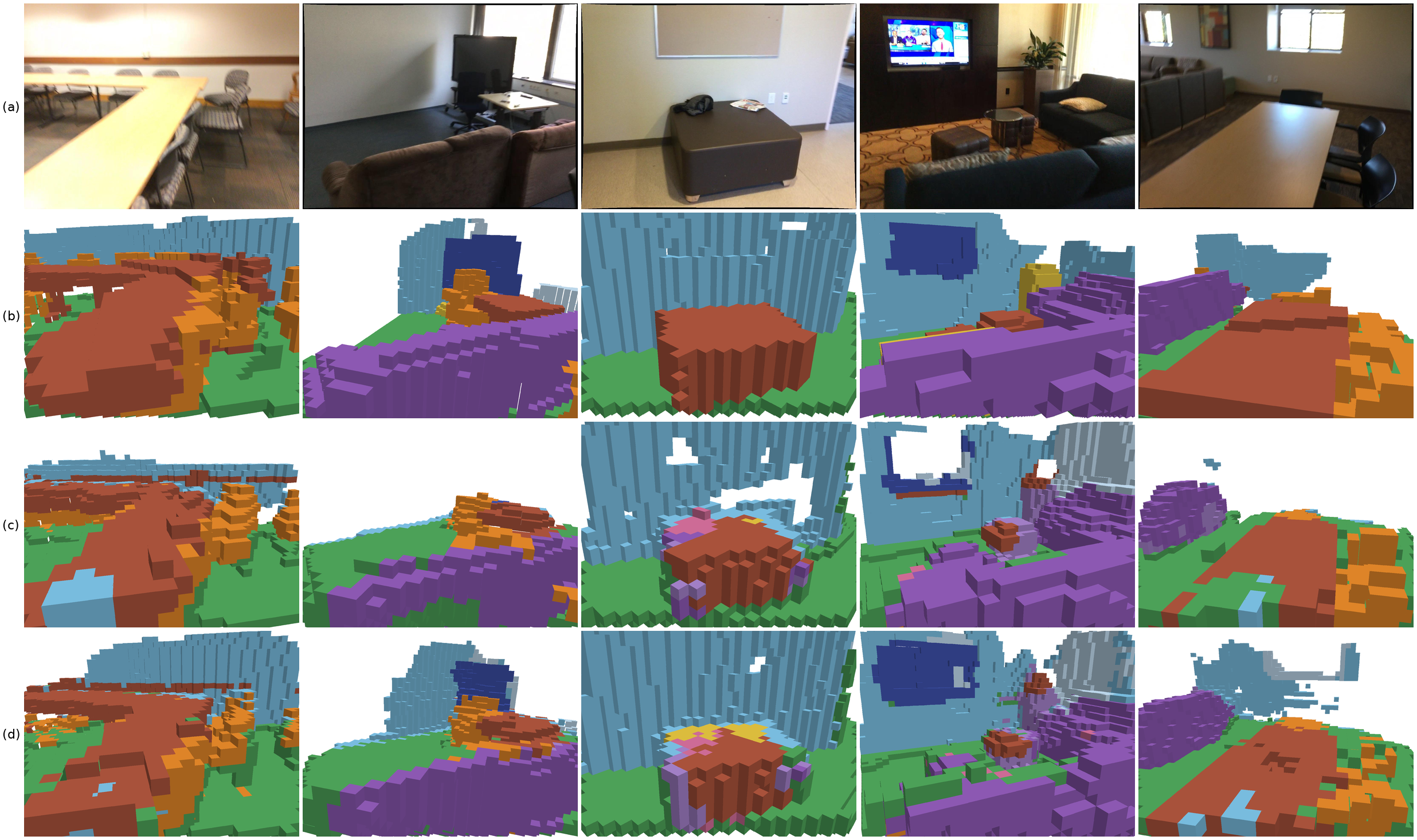}
\caption{Qualitative results on five Occ-ScanNet scenes. From top to bottom: the input photograph, the ground-truth semantic occupancy, the result of the Poisson baseline, and the result of our model.}
\label{fig:qual}
\end{figure*}

\subsection{Main Experiment}
\label{sec:mainexp}

Table~\ref{tab:main} presents the monocular occupancy prediction results on Occ-ScanNet. The closed-vocabulary methods are trained with full semantic annotations, whereas the open-vocabulary methods do not use semantic voxel labels. Under geometry-only supervision, our model achieves 63.31 IoU and 27.44 mIoU, outperforming the strongest compared baseline (59.50 / 21.05) by margins of +3.81 and +6.39, respectively. Notably, significant per-class improvements are observed: \textit{furniture} rises from 5.88 to 26.92 (+21.04), \textit{ceiling} from 17.81 to 35.27 (+17.46), and \textit{bed} from 33.05 to 38.87 (+5.82).

As an extended structural surface, the substantial gain in \textit{ceiling}, along with the overall IoU improvement, is primarily attributed to EC-G2O. Specifically, its mass normalization and $p$-norm aggregation effectively regulate the contributions of overlapping primitives within the same voxel. Conversely, \textit{furniture} and \textit{bed} are discrete object categories that the compared baseline struggles to recognize. The corresponding mIoU gains are driven by the query-conditioned router, which precisely assigns language evidence to individual primitives via MM-QGSR, smooth aggregation, and subclass queries.

Remarkably, our geometric IoU of 63.31 surpasses all closed-vocabulary counterparts, including RoboOcc (56.48), demonstrating that EC-G2O achieves superior occupancy accuracy even without semantic voxel labels. For semantics, our mIoU of 27.44 outperforms closed-vocabulary models like MonoScene (24.62) and TPVFormer (24.94), though it remains slightly below ISO (28.71) and embodied-occupancy models trained with full supervision. This remaining gap reflects the inherent challenge of calibrating fine-grained categories in the absence of dense semantic annotations.

\begin{table}[ht]
\centering
\setlength{\belowcaptionskip}{6pt}
\setlength{\tabcolsep}{3pt}
\caption{Aggregation cost of EC-G2O against the Poisson \mbox{baseline}.}
\label{tab:eff}
\begin{tabular}{@{}lccc@{}}
\toprule
Method &
\begin{tabular}{@{}c@{}}Gaussian-voxel\\pairs\end{tabular} &
\begin{tabular}{@{}c@{}}Aggregation\\latency (ms)\end{tabular} &
\begin{tabular}{@{}c@{}}Full forward\\latency (ms)\end{tabular} \\
\midrule
Baseline & $3.115 \times 10^{6}$ & 19.76 & 60.16 \\
\textbf{EC-G2O} & $\mathbf{1.517 \times 10^{6}}$ & \textbf{12.88} & \textbf{50.88} \\
\bottomrule
\end{tabular}
\end{table}

\paragraph{Efficiency.}
Table~\ref{tab:eff} compares the Poisson baseline with EC-G2O in terms of Gaussian-voxel pairs and overall inference latency, measured on identical hardware. This substantial reduction in evaluated pairs is inherently driven by our geometric aggregation design. By fixing the integrated mass of each primitive to $\alpha_i V$, mass normalization ensures that expanding the spatial scale lowers the local density peak, thereby reducing the number of Gaussians effectively contributing to any given voxel. Geometrically, the $p$-norm taggregation replaces the standard weight $h_i$ with $h_i^{p}$ (where $p>1$). This exponentiation forces contributions to decay more rapidly away from the center, which mathematically tightens the effective culling radius along each semi-axis to $s_{ik}/\sqrt{p}$. Consequently, the total Gaussian-voxel pairs drop drastically from $3.115 \times 10^{6}$ to $1.517 \times 10^{6}$ ($-51.3\%$),  reducing the isolated aggregation time from 19.76 ms to 12.88ms ($-34.8\%$). Since the full forward pass also includes the image backbone and Gaussian head, whose computational costs remain constant, the end-to-end latency achieves a more modest but meaningful reduction from 60.16\,ms to 50.88\,ms ($-15.4\%$).

\subsection{Ablation Study}
\label{sec:ablation}

To validate the contribution of each proposed module, we conduct a cumulative ablation study. Starting from the Poisson baseline, components are progressively incorporated, with the resulting performance detailed in Table~\ref{tab:ablation}.

Introducing mass normalization improves mIoU to 22.82 and IoU to 61.46, indicating that balancing primitive contributions inherently aids semantic recognition. Replacing the standard sum with $p$-norm aggregation drives a substantial geometric leap, pushing IoU to 63.31 (+3.81 over the baseline). However, this modification causes a slight mIoU dip (from 22.82 to 22.64). This trade-off occurs because while $p$-norm aggregation effectively suppresses dominant Gaussians to refine structural boundaries (boosting IoU), it simultaneously attenuates the concentrated feature responses required for distinguishing fine-grained semantics.

\begin{table}[ht]
\centering
\setlength{\belowcaptionskip}{6pt}
\setlength{\tabcolsep}{3pt}
\caption{Cumulative ablation on Occ-ScanNet. Changes from the Poisson baseline are in parentheses.}
\label{tab:ablation}
\begin{tabular}{@{}>{\raggedright\arraybackslash}p{5.5cm}cc@{}}
\toprule
Configuration & IoU & mIoU \\
\midrule
Poisson baseline & 59.50 & 21.05 \\
+ Mass normalization & 61.46 & 22.82 \\
+ $p$-norm aggregation & 63.31 & 22.64 \\
+ MM-QGSR & 63.31 & 23.79 \\
+ Smooth aggregation & 63.31 & 24.73 \\
+ Fine-grained subclass queries & 63.31 & 27.44 \\
\bottomrule
\end{tabular}
\end{table}

The subsequent three modules, namely MM-QGSR, smooth aggregation, and fine-grained subclass queries, are designed explicitly to recover and enhance semantic fidelity. Upon introducing MM-QGSR, mIoU climbs steadily from 22.64 to 23.79, further improves to 24.73 with smooth aggregation, and ultimately surges to 27.44 with fine-grained subclass queries. This final step delivers the most significant semantic boost (+2.71 over smooth aggregation and +6.39 over the baseline), proving its critical role in resolving broad categories. The configuration in the final row corresponds exactly to the full model evaluated in Table~\ref{tab:main}.

\subsection{Qualitative Visualization}
\label{sec:qual}

Figure~\ref{fig:qual} presents qualitative comparisons across five representative scenes, with each column corresponding to a distinct scene. From top to bottom, the four rows display the input RGB image, the ground-truth semantic occupancy, the prediction of the Poisson baseline, and our final result, respectively.

Consistent with the quantitative improvements reported in Table~\ref{tab:main}, compared to the baseline, our method reconstructs more complete \textit{wall} structures, while yielding sharper semantic boundaries for object categories such as \textit{table}, \textit{chair}, and \textit{sofa}. Furthermore, our model predicts \textit{television} instances more completely and assigns their semantic labels with higher accuracy.
\section{Conclusion}
\label{sec:conclusion}
This paper presents GCSR-OCC, a novel framework that advances monocular 3D occupancy prediction via geometric calibration and semantic routing.   Geometrically, our evidence-conserving Gaussian-to-occupancy (EC-G2O) operator effectively suppresses scale inflation and attenuates excessive local evidence accumulation. Semantically, the proposed multimodal query-conditioned Gaussian semantic routing (MM-QGSR) mitigates feature mixing both across and within individual primitives. These designs are further complemented by fine-grained subclass queries and smooth aggregation to enhance open-vocabulary recognition.
Geometry is trained with binary occupancy supervision, and the semantic router is distilled from a frozen 2D teacher. Trained solely without dense 3D semantic annotations, GCSR-OCC achieves a geometric IoU of 63.31\% and an open-vocabulary mIoU of 27.44\% on Occ-ScanNet. Moreover, by substantially reducing Gaussian-voxel pairs during splatting, our method simultaneously decreases aggregation latency, demonstrating that rigorous geometric design can unify prediction accuracy with computational efficiency. Future work will explore generative models for completing the geometry and semantics of occluded regions from a single RGB image.

% \end{document}

% References
{
\small
\bibliographystyle{ieeenat_fullname}
\bibliography{main}
}

% Supplementary material
\clearpage
\clearpage
\setcounter{page}{1}
\maketitlesupplementary
% Supplementary geometry fragment. Included from sec/X_suppl.tex.
\paragraph{Separation of occupancy and semantics.}
EC-G2O maps nonnegative evidence to occupancy by $p_{\mathrm{occ}}=1-\exp(-\gamma z)$. Semantic features $\mathbf{f}_i$ do not enter this mapping. Voxel semantics are produced by MM-QGSR in the main text, not by pooling $\mathbf{f}_i$ on the Poisson support $z=\sum_i h_i$.

\paragraph{Insufficient contribution of one Gaussian.}
A noisy-OR is used by GaussianFormer-2 because peak occupancy of one Gaussian is defined to be $1$, so a single primitive already occupies and overlap is not required. We cannot borrow that remedy by rescaling the peak to $1$: that would abandon the $\alpha_i$ shared with splatting. Under $\alpha_i\le 1$ and an undensified $\varphi_i$, the Poisson intensity writes ``a single Gaussian cannot be high-confidence occupied'' into the representation, and the noisy-OR becomes a reward for copies. EC-G2O is an occupancy aggregator under this constraint, not a different summation chosen in an unconstrained setting.

\paragraph{Scale only grows.}
The renderer still uses the original $\alpha_i$. The scale-dependent factor in the mass-normalized kernel is determined by $\boldsymbol{\Sigma}_i$ and is not a second learnable opacity. Mass normalization already allows the peak of a compact Gaussian to exceed $1$, so a single primitive's $z$ need not be capped at $1$.

\paragraph{Copies raise occupancy.}
The composed occupancy support is $z(\mathbf{x})=\bigl(\sum_i h_i(\mathbf{x})^{p}\bigr)^{1/p}$. The credit assigned to primitive $i$ is
\begin{equation}
\frac{\partial z}{\partial h_i}=\Bigl(\frac{h_i}{z}\Bigr)^{p-1},\qquad p\ge 1.
\end{equation}
At $p=1$ the expression reduces to the original sum, and every overlapping Gaussian receives unit credit regardless of its contribution. At $p>1$, credit concentrates on the dominant primitive. The gradient vanishes for a weak contribution.
% At $p>1$, credit concentrates on the dominant primitive and redundant copies receive vanishing gradient.

\paragraph{Geometry objective.}
Occupancy is supervised on voxels inside the camera frustum. Ignored voxels are excluded. The remaining labels are empty versus occupied, written $y(\mathbf{x})\in\{0,1\}$ below. Let $p_{\mathrm{occ}}=1-\exp(-\gamma z)$ be the EC-G2O occupancy probability. The occupancy loss is
\begin{equation}
\mathcal{L}_{\mathrm{occ}}
=
\lambda_{\mathrm{foc}}\mathcal{L}_{\mathrm{focal}}
+\lambda_{\mathrm{lov}}\mathcal{L}_{\mathrm{lovasz}}
+\lambda_{\mathrm{GS}}\mathcal{L}_{\mathrm{GS}}
+\lambda_{\mathrm{SS}}\mathcal{L}_{\mathrm{SS}},
\label{eq:supp_l_occ}
\end{equation}
applied to $p_{\mathrm{occ}}$ with weights $\lambda_{\mathrm{foc}}=100$ and $\lambda_{\mathrm{lov}}=\lambda_{\mathrm{GS}}=\lambda_{\mathrm{SS}}=1$. Here $\mathcal{L}_{\mathrm{focal}}$ and $\mathcal{L}_{\mathrm{lovasz}}$ are focal classification and Lov\'asz-softmax, and $\mathcal{L}_{\mathrm{GS}}$, $\mathcal{L}_{\mathrm{SS}}$ are geometric and semantic scene-scale terms on empty versus occupied. These four terms follow the standard voxel-occupancy recipe and are not the geometric contribution of EC-G2O.

Because $\partial p_{\mathrm{occ}}/\partial z=\gamma\exp(-\gamma z)$ vanishes once $z$ is large, $\mathcal{L}_{\mathrm{occ}}$ barely penalizes further stacking. The two intensity losses $\mathcal{L}_{\log z}$ and $\mathcal{L}_{\mathrm{over}}$ in the main-text objective are applied to $z$, with masks taken from ground truth rather than from the prediction. Let $z^{\ast}=-\log(1-0.95)\approx 3$. The log-intensity term is a Huber regression of $\log(1+z)$:
\begin{equation}
\mathcal{L}_{\log z}
=
\mathrm{Huber}\!\bigl(\log(1+z),\, t(y)\bigr),
t(1)=\log(1+z^{\ast}),
t(0)=0.
\label{eq:supp_logz}
\end{equation}
Its gradient $1/(1+z)$ remains much larger than $\partial p_{\mathrm{occ}}/\partial z$ in the saturated regime, so surplus evidence on empty voxels is driven toward $0$ and $z$ on occupied voxels is anchored at a finite scale. The overshoot term applies only on ground-truth occupied voxels:
\begin{equation}
\mathcal{L}_{\mathrm{over}}
=
\mathbb{E}\bigl[\mathrm{relu}(z-z^{\ast})^{2}\bigr].
\label{eq:supp_overshoot}
\end{equation}
Values below $z^{\ast}$ are not penalized; overshoot is penalized quadratically. The geometry objective used in training is
\begin{equation}
\mathcal{L}_{\mathrm{geo}}
=
\mathcal{L}_{\mathrm{occ}}
+\lambda_{\log z}\mathcal{L}_{\log z}
+\lambda_{\mathrm{over}}\mathcal{L}_{\mathrm{over}},
\label{eq:supp_l_geo}
\end{equation}
with $\lambda_{\log z}=0.5$ and $\lambda_{\mathrm{over}}=0.05$. An image-space depth Huber loss (valid pixels only) and a frozen 2D feature-alignment term are used as implementation auxiliaries; neither enters $p_{\mathrm{occ}}$, and occupancy aggregation does not use the semantic features $\mathbf{f}_i$.

\paragraph{Semantic objective.}
Binary occupancy supervision does not train $W_g$, $W_t$, $\eta$, $\lambda_m$, $\lambda_v$, or the mode parameters, because occupancy does not depend on the semantic features. These parameters are trained by distilling the frozen Trident teacher. The teacher is not updated, and no semantic voxel label is used. Gradients enter only the router.

Two losses are applied. The image-level loss has weight $0.5$. Query scores are rendered into the image and bilinearly resampled to the resolution of the teacher feature map. Their softmax is matched, by a KL divergence, to the softmax of the frozen dense features dotted with the CLIP text embeddings. A relation term of weight $0.25$ matches the mean-centered teacher and student logits by cosine distance. Both terms are weighted by rendered opacity and teacher confidence. The teacher and student logit scales are $20$.

The footprint loss has weight $1.0$. In each training view, the Gaussian covariance is projected to an ellipse of radius $1.5$ standard deviations and sampled at the center and on one ring of eight points. Frozen teacher features at these locations remain separate observations and are not averaged into one target. A soft mixture of temperature $0.25$ assigns each observation to the latent mode that explains it. A diversity term of weight $0.1$ and a mode-mass term of weight $0.05$ keep the two modes from collapsing.

For each Gaussian and query, we first combine modes into three evidence moments:
\begin{equation}
b_{ic}^{(j)}=\sum_{k=1}^{K}r_{ikc}e_{ikc}^{j},
\qquad j\in\{0,1,2\}.
\label{eq:supp_mm_packed_moments}
\end{equation}
The zeroth, first, and second moments encode routing support, weighted text matching, and weighted squared text matching, respectively. They are spatially aggregated for voxel inference and rendered for image supervision:
\begin{equation}
\begin{aligned}
A_c^{(j)}(\mathbf x)&=\sum_i\bar g_i(\mathbf x)b_{ic}^{(j)},\\
B_{uc}^{(j)}&=\sum_i a_{ui}b_{ic}^{(j)},
\end{aligned}
\label{eq:supp_mm_spatial_moments}
\end{equation}
where $\bar g_i(\mathbf x)$ are the normalized semantic spatial weights and $a_{ui}$ are the rasterizer's alpha-compositing coefficients at pixel $u$. The image moments are not additionally divided by rendered opacity. For a moment triplet $T=(T^{(0)},T^{(1)},T^{(2)})$, both paths use the decoder
\begin{equation}
\begin{aligned}
\widetilde m(T)&=\max(T^{(0)},\epsilon),\\
\mu(T)&=T^{(1)}/\widetilde m(T),\\
v(T)&=[T^{(2)}/\widetilde m(T)-\mu(T)^2]_+,\\
\mathcal D(T)&=\mu(T)+\lambda_m\frac{\log\widetilde m(T)}{\beta}
-\lambda_vv(T).
\end{aligned}
\label{eq:supp_mm_decoder}
\end{equation}
Here $[x]_+=\max(x,0)$. Thus, voxel scores are $S_c(\mathbf x)=\mathcal D(A_c(\mathbf x))$, and rendered pixel scores are $s_{uc}=\mathcal D(B_{uc})$. The two paths share the decoder but use different spatial coefficients. This order---mode moments, rendering, and score decoding---allows image supervision to update the mode, routing, and score-calibration parameters.

\paragraph{Implementation.}
The identity below is used to evaluate $h_i^{p}$. After mass normalization, a primitive contribution factors as the unnormalized kernel times a scalar that depends only on the primitive itself. That scalar is the coefficient in front of the Gaussian exponential, times $\alpha_i V$, and does not depend on the query $\mathbf{x}$, so it can be absorbed into opacity. The effective opacity is
\begin{equation}
\tilde{\alpha}_i=\frac{\alpha_i V}{(2\pi)^{3/2}s_{i1}s_{i2}s_{i3}},
\qquad
h_i(\mathbf{x})=\tilde{\alpha}_i\,\varphi_i(\mathbf{x}).
\end{equation}
Here $\tilde{\alpha}_i$ folds the voxel volume and the three semi-axes into opacity, while the shape at the query remains $\varphi_i$. Because $\varphi_i$ is unnormalized, $\varphi_i(\boldsymbol{\mu}_i)=1$, and raising it to the $p$-th power acts only on the exponent:
\begin{equation}
\begin{aligned}
\varphi_i(\mathbf{x})^{p}
&=\exp\Big(-\tfrac{p}{2}(\mathbf{x}-\boldsymbol{\mu}_i)^{\top}\boldsymbol{\Sigma}_i^{-1}(\mathbf{x}-\boldsymbol{\mu}_i)\Big)\\
&=\exp\Big(-\tfrac12(\mathbf{x}-\boldsymbol{\mu}_i)^{\top}(p\boldsymbol{\Sigma}_i^{-1})(\mathbf{x}-\boldsymbol{\mu}_i)\Big).
\end{aligned}
\end{equation}
The right-hand side has the same form as $\varphi_i$, with precision $\boldsymbol{\Sigma}_i^{-1}$ replaced by $p\boldsymbol{\Sigma}_i^{-1}$ (equivalently, covariance $\boldsymbol{\Sigma}_i/p$). Therefore
\begin{equation}
h_i(\mathbf{x})^{p}
=\tilde{\alpha}_i^{p}\,
\exp\Big(-\tfrac12(\mathbf{x}-\boldsymbol{\mu}_i)^{\top}(p\boldsymbol{\Sigma}_i^{-1})(\mathbf{x}-\boldsymbol{\mu}_i)\Big),
\end{equation}
which is the same kernel evaluated on a primitive with parameters $(\tilde{\alpha}_i^{p},\,\boldsymbol{\Sigma}_i/p)$. The occupancy field is therefore
\begin{equation}
z(\mathbf{x})
=\left(\sum_{i=1}^{N}
\tilde{\alpha}_i^{p}\,
\exp\Big(-\tfrac12(\mathbf{x}-\boldsymbol{\mu}_i)^{\top}(p\boldsymbol{\Sigma}_i^{-1})(\mathbf{x}-\boldsymbol{\mu}_i)\Big)
\right)^{1/p},
\end{equation}
and occupancy is $p_{\mathrm{occ}}=1-\exp(-\gamma z)$.

\paragraph{Experiment Details.}
We use Depth Anything V2 (ViT-B) as the image backbone. Its features are converted into 3D primitives by a Gaussian head and projected onto the voxel grid by EC-G2O. Category names are embedded by a frozen CLIP ViT-B/16 text encoder. Dense language-aligned features come from a frozen Trident segmenter (CLIP ViT-B/16, DINO, and SAM). Input images are resized so that the longer side is 518 pixels, preserving the aspect ratio.

Geometry is optimized with AdamW (learning rate $2\times 10^{-4}$, weight decay $0.01$, gradient clipping $1.0$), a cosine schedule, and linear warmup. The backbone learning rate is scaled by $0.1$. The geometry branch is trained for 15 epochs. The splatting temperature decays exponentially from $1$ to $10^{-3}$ over the first 10 epochs and is then held constant. We set $p=2$. The per-GPU batch size is 1, using three NVIDIA RTX 4090 GPUs.

The router is trained for two further epochs after occupancy parameters are frozen, with learning rate $1\times 10^{-4}$. Each primitive has $K=2$ semantic modes. At inference, semantic aggregation uses $q=0.5$ and the subclass queries defined below. Evaluation uses voxels inside the camera frustum. Geometric accuracy is the IoU between predicted and ground-truth occupied voxels. Semantic accuracy is the per-class IoU of the eleven nonempty categories and their mean (mIoU). The empty class is excluded from mIoU.

\paragraph{Subclass queries.}
Only the hypernyms \textit{furniture} and \textit{objects} are expanded. For a parent word, the language-model prompt is: ``List the 20 most common concrete object types that would be described by the word `\{parent\}' in an ordinary indoor room. Output only a comma-separated list of singular noun phrases, ordered from most common to least common.'' The prompt names neither the dataset nor the sibling classes.

%%%%%%%%%%%%%%%%%%%%%%%%%%%%%%%%%
% 语义部分的支撑材料，多模式的详细部分
\paragraph{Multi-mode semantic parameterization.}
Let $\mathbf f_i\in\mathbb R^{512}$ be the normalized, frozen feature of Gaussian $i$, and let $\mathbf t_c\in\mathbb R^{512}$ be the normalized text embedding of query $c$. Two independent MLPs generate residual mode features and mode priors:
\begin{align}
\mathbf f_{ik}
&=\operatorname{norm}\!\left(\mathbf f_i+\rho\tanh([F_\theta(\mathbf f_i)]_k)\right),
\label{eq:supp_mm_features}\\
\pi_{ik}
&=\operatorname{softmax}_k(P_\psi(\mathbf f_i)).
\label{eq:supp_mm_priors}
\end{align}
Here, $\operatorname{norm}$ denotes $\ell_2$ normalization, and $[F_\theta(\mathbf f_i)]_k$ is the $k$-th row after reshaping the output into $K\times512$. We use $K=2$ and $\rho=0.25$. The residual and prior MLPs have widths $512\!\to\!128\!\to\!512K$ and $512\!\to\!128\!\to\!K$, respectively, with GELU between their linear layers. The final residual layer is initialized with zero bias and Gaussian weights of standard deviation $10^{-3}$. The final prior layer is initialized to zero, giving uniform initial priors; earlier layers use the default linear-layer initialization. Parameters inherited from earlier training stages retain their checkpoint values. All modes of a primitive share its spatial kernel.

\paragraph{Text-conditioned routing.}
Mode-level evidence, alignment logits, and routing weights are
\begin{equation}
\begin{aligned}
e_{ikc}&=\mathbf f_{ik}^{\top}\mathbf t_c,\\
\ell_{ikc}&=e_{ikc}+\tanh(\eta)\operatorname{cos\_sim}
(W_g\mathbf f_{ik},W_t\mathbf t_c),\\
r_{ikc}&=\pi_{ik}\exp[\operatorname{clip}(\beta\ell_{ikc},-L,L)].
\end{aligned}
\label{eq:supp_mm_routing}
\end{equation}
The bias-free projections $W_g,W_t$ map 512-dimensional features to 64 dimensions and use Xavier initialization; $\eta$ is initialized to zero. Routing sharpness and score-calibration parameters are constrained by
\begin{equation}
\begin{aligned}
\beta&=\frac{0.25+3.75\operatorname{sigmoid}(b_\beta)}{0.25},\\
\lambda_m&=\operatorname{softplus}(b_m),\qquad
\lambda_v=\operatorname{softplus}(b_v).
\end{aligned}
\label{eq:supp_mm_calibration}
\end{equation}
Thus $\beta\in(1,16)$ and $\lambda_m,\lambda_v\geq0$. Configured initial values are $\beta=4$ and $\lambda_m=\lambda_v=0.25$. We set $L=5$ and use $\epsilon=10^{-6}$ in the stabilized expressions below.

\paragraph{Training queries.}
We use $C=64$ fixed text queries. Each query is encoded with the \texttt{openai\_imagenet\_template} prompts: individual CLIP embeddings are normalized, averaged, and normalized again. The distillation queries are listed below; they exclude the eleven benchmark parent strings. This exclusion does not imply that all subclass queries used at evaluation are unseen.
\begin{quote}\small
door, doorway, pillar, column, beam, staircase, railing, balcony, roof, carpet, rug, curtain, blind, molding, cabinet, desk, shelf, bookshelf, dresser, nightstand, wardrobe, stool, bench, counter, closet, cupboard, drawer, ottoman, armchair, workstation, rack, box, book, bag, lamp, pillow, bottle, cup, towel, clothing, plant, trash bin, monitor, keyboard, picture frame, cushion, basket, backpack, vase, telephone, clock, refrigerator, microwave, oven, sink, bathtub, toilet, mirror, fan, heater, air conditioner, laptop, printer, speaker.
\end{quote}

\paragraph{Subclass query mapping.}
Table~\ref{tab:subclass_query_mapping} lists the retained subclasses for \textit{furniture} and \textit{objects}. The other nine parent queries (ceiling, floor, wall, window, chair, bed, sofa, table, and television) are used without expansion. Each parent score is the maximum over its own query score and the scores of its retained subclasses.

\begin{table}[H]
\centering
\caption{Retained subclass queries. All 11 parent queries remain in the vocabulary, giving 29 queries in total.}
\label{tab:subclass_query_mapping}
\small
\setlength{\tabcolsep}{4pt}
\renewcommand{\arraystretch}{1.12}
\begin{tabular}{@{}lc>{\raggedright\arraybackslash}p{0.60\columnwidth}@{}}
\toprule
\textbf{Parent} & \textbf{Count} & \textbf{Retained subclass queries} \\
\midrule
furniture & 6 & cabinet, bookshelf, dresser, chest of drawers, ottoman, filing cabinet \\
\addlinespace[3pt]
objects & 12 & box, book, bag, lamp, bottle, cup, clothes, potted plant, trash can, basket, backpack, vase \\
\bottomrule
\end{tabular}
\end{table}

\paragraph{Subclass queries on the Poisson baseline.}
To isolate the contribution of fine-grained prompts, we apply the identical set of retained subclass queries to the Poisson baseline (denoted as LegoOcc in Table~\ref{tab:main}), with results detailed in Table~\ref{tab:subclass_control}. Since this modification solely replaces the text prototypes without altering the underlying geometry, the occupancy field remains intact, keeping the geometric IoU fixed at $59.50$. mIoU rises from $21.05$ to $24.63$. Our full model still reaches $27.44$ mIoU, a further $+2.81$ over this prompt-matched baseline. Subclass queries therefore explain part of the semantic gain, but not all of it: EC-G2O, MM-QGSR, and smoothed aggregation contribute the remaining mIoU once the query set is held fixed.

\begin{table}[ht]
\centering
\setlength{\belowcaptionskip}{6pt}
\setlength{\tabcolsep}{3pt}
\caption{Subclass queries on the Poisson baseline. The first row repeats the LegoOcc numbers from Table~\ref{tab:main}. The second row uses the same retained queries as our full model.}
\label{tab:subclass_control}
\begin{tabular}{@{}lcc@{}}
\toprule
Configuration & IoU & mIoU \\
\midrule
LegoOcc & 59.50 & 21.05 \\
+ Subclass queries & 59.50 & 24.63 \\
\textbf{Ours} & \textbf{63.31} & \textbf{27.44} \\
\bottomrule
\end{tabular}
\end{table}

\paragraph{Gaussian scale.}
By fixing the integrated mass of each primitive to $\alpha_i V$, mass normalization in EC-G2O ensures that enlarging a Gaussian inherently lowers its local density peak, thereby biasing the occupancy objective toward compact primitives. To empirically validate this effect, Table~\ref{tab:gaussian_scale} reports the Gaussian scales—defined as the mean of the three semi-axes—for active primitives participating in occupancy aggregation across the first frames of $100$ validation scenes. Compared to the Poisson baseline, EC-G2O significantly reduces both the mean scale (from $0.0479\,\mathrm{m}$ to $0.0349\,\mathrm{m}$, $-27.1\%$) and the median scale (from $0.0458\,\mathrm{m}$ to $0.0341\,\mathrm{m}$, $-25.4\%$). This contracted spatial footprint confirms that the learned Gaussians naturally favor smaller volumes, perfectly aligning with the theoretical scale-peak trade-off derived in our method.

\begin{table}[ht]
\centering
\setlength{\belowcaptionskip}{6pt}
\setlength{\tabcolsep}{3pt}
\caption{Gaussian scale on $100$ validation scenes. Scale is the mean of the three semi-axes, in metres.}
\label{tab:gaussian_scale}
\begin{tabular}{@{}lcc@{}}
\toprule
Method & Mean (m) & Median (m) \\
\midrule
Poisson baseline & 0.0479 & 0.0458 \\
\textbf{EC-G2O} & \textbf{0.0349} & \textbf{0.0341} \\
\bottomrule
\end{tabular}
\end{table}

\end{document}